\documentclass[10pt,twocolumn,letterpaper]{article}
\usepackage{algorithm}
\usepackage{algpseudocode}
 \usepackage{multirow}
\usepackage{cvpr}              % To produce the CAMERA-READY version
\usepackage{times}
\usepackage{epsfig}
\usepackage{graphicx}
\usepackage{amsmath}
\usepackage{amssymb}

\usepackage{multirow}
\usepackage{bbm}
\usepackage{cuted}

\definecolor{cvprblue}{rgb}{0.21,0.49,0.74}
\usepackage[pagebackref,breaklinks,colorlinks,allcolors=cvprblue]{hyperref}

\def\paperID{9256} % *** Enter the Paper ID here
\def\confName{CVPR}
\def\confYear{2025}

\title{AMR-Transformer: Enabling Efficient Long-range Interaction for Complex Neural Fluid Simulation}

\author{
    Zeyi Xu\textsuperscript{$2$}\thanks{Co-first authors.}~~~~
    Jinfan Liu\textsuperscript{$1$}\footnotemark[1]~~~~
    Kuangxu Chen\textsuperscript{$3$}~~~~
    Ye Chen\textsuperscript{$1$}~~~~
    Zhangli Hu\textsuperscript{$1$}~~~~
    Bingbing Ni\textsuperscript{$1$}\thanks{Corresponding author.}\\
    $^1$Shanghai Jiao Tong University~~~~
    $^2$Shanghai University~~~~
    $^3$Shenzhen Technology University\\
    \tt\small xzyblxa@shu.edu.cn~~~\{ljf\_{2024}, nibingbing\}@sjtu.edu.cn\\
    \href{https://github.com/JfanLiu/AMR_Transformer}{\texttt{https://github.com/JfanLiu/AMR\_Transformer}}
}

\begin{document}
\maketitle

\begin{strip}
    \centering
    \includegraphics[width=0.99\textwidth]{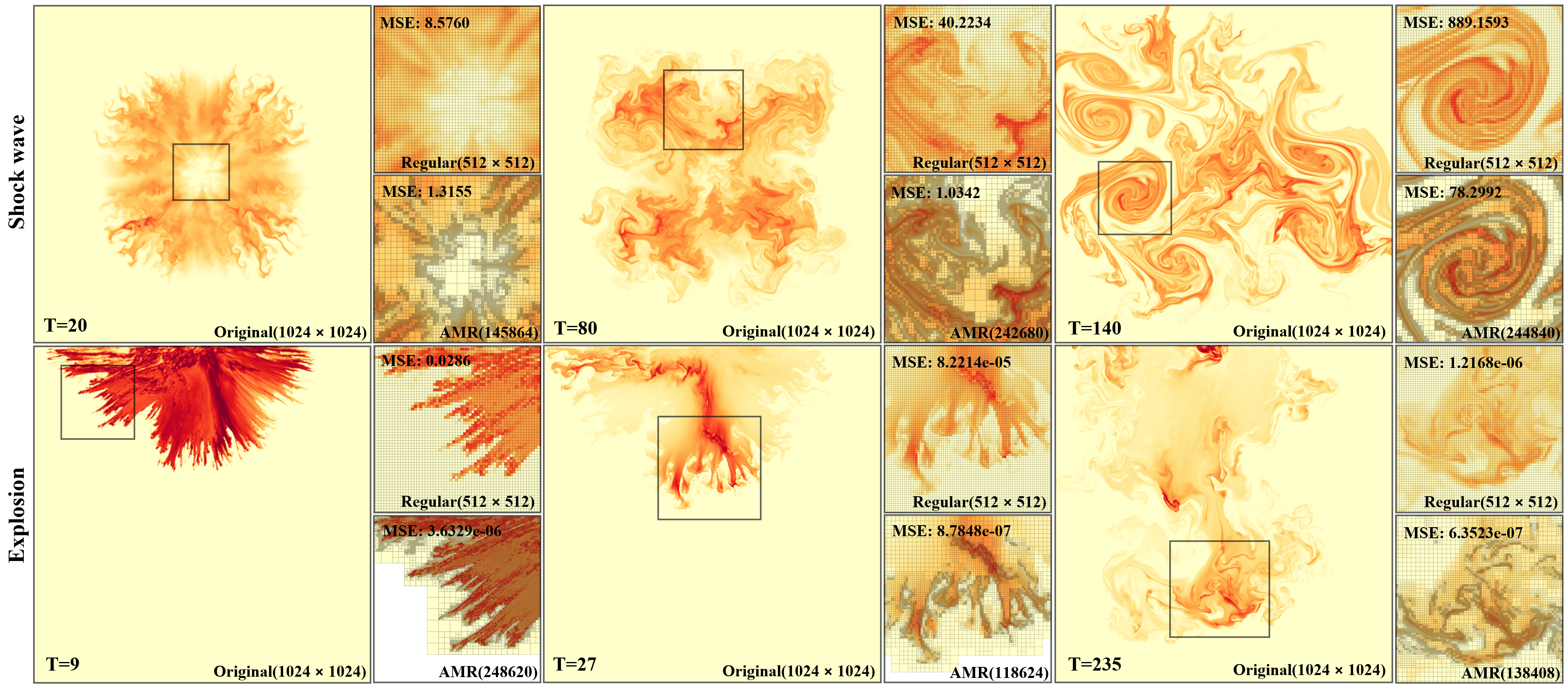}
    \captionof{figure}{
    \textbf{AMR tokenizer results in high-resolution simulation.} Visualization of the AMR tokenizer applied to 1024 × 1024 shock wave and explosion simulations. The AMR tokenizer captures fine-scale structures while reducing token count. Right panels compare the regular 512 × 512 grid with the AMR partitioning, each right panel displays the total cell count and partitioning scheme (bottom-right), along with the mean squared error (MSE) relative to the 1024 × 1024 ground truth (top-left).
    }
    \label{fig:amr_tokenizer_results}
% \vspace{-0.4cm}
\end{strip}

\vspace{-0.3cm}
\begin{abstract}
% \vspace{-0.3cm}
Accurately and efficiently simulating complex fluid dynamics is a challenging task that has traditionally relied on computationally intensive methods. Neural network-based approaches, such as convolutional and graph neural networks, have partially alleviated this burden by enabling efficient local feature extraction. However, they struggle to capture long-range dependencies due to limited receptive fields, and Transformer-based models, while providing global context, incur prohibitive computational costs. To tackle these challenges, we propose AMR-Transformer, an efficient and accurate neural CFD-solving pipeline that integrates a novel adaptive mesh refinement scheme with a Navier-Stokes constraint-aware fast pruning module. This design encourages long-range interactions between simulation cells and facilitates the modeling of global fluid wave patterns, such as turbulence and shockwaves. Experiments show that our approach achieves significant gains in efficiency while preserving critical details, making it suitable for high-resolution physical simulations with long-range dependencies. On CFDBench, PDEBench and a new shockwave dataset, our pipeline demonstrates up to an order-of-magnitude improvement in accuracy over baseline models. Additionally, compared to ViT, our approach achieves a reduction in FLOPs of up to 60 times.
\end{abstract}
\vspace{-1cm}

% Accurately and efficiently simulating complex fluid dynamics is a challenging task that has traditionally relied on computationally intensive methods. Neural network-based approaches, such as convolutional and graph neural networks, have partially alleviated the burden of these classical methods by providing efficient local feature extraction. However, they struggle with capturing long-range dependencies due to limited receptive fields, and Transformer-based models, while providing global context, incur prohibitive computational costs. To address these challenges, we propose AMR-Transformer, an efficient and accurate neural CFD solving pipeline that integrates a novel adaptive mesh refinement scheme with a Navier-Stokes constraint-aware fast pruning module. This design encourages long-range interactions between simulation cells and facilitates global fluid wave pattern modeling, such as turbulence and shockwaves. Experiments show that our pipeline demonstrates up to an order-of-magnitude improvement in accuracy over baseline models. Additionally, compared to ViT, our approach achieves up to an order-of-magnitude reduction in FLOPs.

% 由于高计算成本和需要捕获长期依赖关系，准确有效地模拟复杂的流体动力学具有挑战性。传统的模型，如cnn，由于其有限的接受域，经常在高分辨率模拟中挣扎，而变形金刚以令人望而却步的计算费用提供全局背景。为了解决这个问题，我们引入了AMR-Transformer，它结合了自适应网格细化（AMR）和Transformer神经求解器。AMR组件根据物理标准自适应地细化区域，通过将资源集中在复杂区域来减少令牌计数。实验表明我们的方法在保留关键细节的同时实现了显著的效率提升，使其适用于高分辨率物理模拟等场景。在CFDBench和提供的一个新的激波数据集下，我们的pipeline相比baselines精度最高提升了一个数量级。并且相比ViT ，FLOPs最高下降了一个数量级。
\vspace{+0.3cm}
\section{Introduction}
\label{sec:intro}
\vspace{-0.1cm}
% 可能需要删减一些
The use of artificial intelligence methods for solving partial differential equations (PDEs) has become increasingly prevalent in computational fluid dynamics (CFD), a field with critical applications in science and engineering, including turbulence modeling, airfoil design, and micromechanics. Traditional CFD simulations are computationally intensive, particularly when modeling complex fluid behaviors like turbulence and vortex dynamics. Deep learning methods provide faster, data-driven approximations of these complex physical processes.

To model local interaction relationships, pioneer methods typically utilize computational schemes like convolutional neural networks (CNNs)~\cite{fno} and graph neural networks (GNNs)~\cite{mpt,mgn} . These methods can be combined with techniques like Level of Detail (LOD)~\cite{neural-flow-map} or multiscale frameworks~\cite{unet} to expand the receptive field. They have been successfully applied to simulate fluid dynamics phenomena governed by Navier-Stokes equations, such as cylinder flow and airfoil dynamics~\cite{msmgn}. In contrast, implicit neural representations (INRs)~\cite{deeponet,pinn} focus on learning direct mappings from input to output, bypassing the explicit modeling of any interactions. These approaches are primarily suitable for single-instance solutions under specific conditions and exhibit limited generalization across different physical instances.

In scenarios that require modeling long-range dependencies, such as high-precision shockwave simulations, traditional local operations may face challenges in capturing the inherent global dependencies. While CNNs, especially architectures like UNets, are capable of global communication through hierarchical structures (such as skip connections), they still rely on a fixed receptive field that can limit their ability to model extremely large-scale global interactions. On the other hand, Transformers, through self-attention mechanisms, can directly model global interactions across the entire computational domain, without the limitations of a fixed receptive field. However, for an input sequence or grid of size \(N\), self-attention has a computational complexity of \(O(N^2)\). High-precision fluid dynamics simulations, for instance, can involve millions or even billions of grid points per time step, leading to unaffordable computational costs and memory requirements for self-attention mechanisms.

Furthermore, in physical simulations, different regions contribute varying levels of information and have different impacts on the overall dynamics. Some areas require high-resolution local modeling to capture complex phenomena accurately, while others do not need the same level of detail. Treating all regions equally is inefficient and can lead to significant redundancy, especially in simulations with heterogeneous complexities. While patch-based methods like Vision Transformer (ViT)~\cite{vit} attempt to address computational costs by dividing the input into patches, ViT treats all patches indiscriminately, lacking tailored adjustments for regions of interest. This limitation restricts the ability to capture critical details in complex physical simulations.

Given these limitations, there is a clear need for a method that \textbf{efficiently models global interactions without incurring excessive computational costs}. To address this, we propose the \textbf{AMR-Transformer} pipeline, which combines Adaptive Mesh Refinement (AMR) as a tokenizer with an Encoder-only Transformer neural solver to model long-range dependencies. Our approach employs a hierarchical multi-way tree-based AMR structure, merging redundant information while preserving multi-scale features. In addition, a Navier-Stokes constraint-aware fast pruning module based on physical properties such as velocity gradient, vorticity, momentum, and phenomena like the Kelvin-Helmholtz instability, guides adaptive focus on regions with complex dynamics. During training, we randomly generate hyperparameters for the subdivision criteria, allowing for manual adjustment post-training to optimize the balance between accuracy and efficiency.

Extensive experiments are conducted and analyzed on the \texttt{CFDBench}~\cite{cfdbench} and \texttt{PDEbench}~\cite{PDEBENCH} benchmarks, which include incompressible flow and PDE-based problems, to evaluate the framework comprehensively. Additionally, we provide a new shockwave dataset with four times the resolution of the \texttt{CFDBench} dataset, introducing compressible flow scenarios, that feature a more uneven distribution of physical information and significantly finer-grained details. Experiments show that our pipeline achieves state-of-the-art (SOTA) performance in most problems, with accuracy improvements of up to an order of magnitude compared over previous SOTA. Our tokenizer can effectively reduce the number of tokens by a factor of 2 to 10 across various problems, with negligible loss in simulation accuracy. Given the \(O(N^2)\) complexity of self-attention, the reduced token count translates into a substantial speedup, approximately quadratic in relation to the token reduction.

% Our main contributions are as follows:
% \begin{enumerate}
%     \item \textbf{New AMR-Transformer Pipeline:} We propose a novel AMR-Transformer pipeline that integrates adaptive mesh refinement (AMR) with a Transformer-based neural solver, effectively balancing computational efficiency with the need for capturing complex fluid dynamics interactions. The pipeline allows AMR refinement parameters to be adjusted post-training, enabling fine-tuning of the tradeoff between accuracy and computational efficiency.
    
%     \item \textbf{Nested multi-way tree based AMR:} We introduce a Nested multi-way tree based AMR technique, which combines the efficient adaptive refinement capabilities of multi-way tree structures with the multi-layer information retention properties of nested grids, enhancing both spatial resolution and data efficiency in dynamic regions. The Navier-Stokes constraint-aware fast pruning module

%     \item \textbf{High-Resolution Shockwave Dataset:} We provide a high-resolution shockwave dataset, which features a highly uneven distribution of physical information, offering a challenging benchmark for evaluating high-precision fluid simulations.
% \end{enumerate}
\vspace{-0.1cm}
\section{Related Work}
\label{sec:related_work}
\subsection{AI for CFD}
Recent advancements in artificial intelligence show great promise in solving CFD problems, providing faster and more efficient alternatives to traditional methods. Physics-Informed Neural Networks (PINNs)~\cite{pinns-b, pinns-wave, pinn} integrate physical constraints directly into the loss function, ensuring compliance with governing equations. DeepONet~\cite{deeponet, DeepONet-multi, s-deeponet, DeepONet-long-time} leverages a fully connected network to learn nonlinear operators, demonstrating strong performance on small datasets.
The Fourier Neural Operators (FNOs)~\cite{fno, fno-factorized, U-FNO} combine convolutional networks with fast Fourier transforms (FFT) for efficient PDE solving. 
% Due to these methods primarily capture local dependencies, they face challenges with high-frequency physical variations and nonlinearity. 
Graph-based neural networks (GNNs)~\cite{gnn,gnn-large}, which model simulation domains as graphs, have led to advancements like the Graph Neural Simulator (GNS)~\cite{gns,Constraint-gns} and Graph Neural Operator (GNO)~\cite{gno,ugrapher-gno}. These methods primarily capture local dependencies, which can lead to insufficient accuracy.
% , enabling fluid simulations that outperform traditional methods like FEM. 
Transformer-based PDE solvers~\cite{transformers-geometries,transformers-muti,transformers-reduce,transformers-scalable,transformers-unisolver} partition simulation domains into token sequences, capturing complex physical correlations while efficiently handling large, high-dimensional spaces. Despite they exhibit both local and global adaptability through attention mechanisms, Transformers require substantial computational resources, especially for large-scale simulations.

\subsection{Accumulation Approaches}
Recent advancements in accumulation techniques for PDE solvers improve computational efficiency, memory usage, and accuracy in high-resolution simulations. 
MultiScale MeshGraphNet~\cite{meshgraphnet-adaptive,meshgraphnet-edgeaggregation,meshgraphnet-stride} employs a multi-scale structure to facilitate coarse-to-fine dynamics learning, thereby enhancing its ability to extract meaningful information across varying resolutions. However, the approach has limitations in high-dimensional scenarios requiring fine-grained accuracy, such as turbulent flow, due to the inherent loss of detailed information. Comparable methodologies include Multi-Grid Neural Operators~\cite{multi-grid-adaptive,multi-grid-geometric,multi-grid-Physics-informed}.
Neural Flow Maps~\cite{neural-flow-map,neural-flow-map-shallow} achieve high accuracy in fluid simulations using bidirectional flow maps, but their effectiveness is constrained by spatial sparsity.
Fourier PINN~\cite{fourier-pinn,fourier-pinn-boundary} improves computational efficiency by replacing costly differential operators with spectral multiplications, while MG-TFNO~\cite{multi-grid-fno} employs Fourier-domain tensor factorization to achieve parameter compression and enable parallelization. However, these methods exhibit limitations in handling irregular or non-uniform simulation domains, as well as in scenarios where spectral approaches are less effective.
\vspace{-0.2cm}

% Fourier PINN~\cite{fourier-pinn,fourier-pinn-boundary} tackles the limitations of traditional Physics-Informed Neural Networks (PINNs) by replacing costly differential operators with spectral multiplications, reducing memory and training time. MultiScale MeshGraphNet~\cite{meshgraphnet-adaptive,meshgraphnet-edgeaggregation,meshgraphnet-stride} enhances MeshGraphNets~\cite{meshgraphnet} with a multi-scale structure that enables coarse-to-fine dynamics learning, alleviating message-passing bottlenecks in high resolutions. Neural Flow Maps~\cite{neural-flow-map,neural-flow-map-shallow} use spatially sparse neural fields for accurate velocity field representation, achieving high fidelity in fluid simulations through bidirectional flow maps with minimal dissipation. Multi-grid Neural Operators~\cite{multi-grid-adaptive,multi-grid-geometric,multi-grid-Physics-informed} address memory constraints and scalability by training on progressively coarser grids, accelerating simulations while ensuring accuracy. Extending the multi-grid concept, MG-TFNO~\cite{multi-grid-fno} incorporates Fourier-domain tensor factorization for parameter compression and parallelism, ideal for turbulent Navier-Stokes simulations. 
% These approaches advance efficient and scalable operator learning, enabling accurate, high-resolution PDE simulations across diverse physical systems.
% \vspace{-0.1cm}
\section{Methodology}
\label{sec:method}
Traditional methods, such as CNN-based approaches, are unable to model long-range interactions effectively, while using Transformers with self-attention to model direct interactions is too slow when handling high-resolution data. In order to alleviate the problem, we propose a novel pipeline, the \textbf{AMR-Transformer}, which combines Adaptive Mesh Refinement (AMR) as a tokenizer with a Transformer-based neural solver. Grid-based AMR processing transforms the feature domain from the structured grid \(\mathbb{R}^{H \times W \times c}\), where each pixel has \(c\)-dimensional features, to a patch representation \(\mathbb{R}^{N \times K \times c}\). Here, \(N\) denotes the number of patches, each patch contains \(K\) cells. For adaptive refinement, we employ a multi-way tree structure, and the hierarchical AMR produces overlapping patches of varying sizes. 
% This transformation allows the Transformer to take patches directly as inputs, where the self-attention mechanism models interactions across multiple scales within the adaptive resolution provided by AMR.

\subsection{AMR Tokenizer}
\label{sec:tokenizer}
The AMR Tokenizer utilizes an adaptive mesh refinement (AMR) strategy based on a hierarchical  multi-way tree structure to capture multi-scale features in fluid dynamics. Refinement decisions are made solely based on the velocity field within the input features, without requiring additional specific input. For simplicity, we illustrate this process in 2D using a quadtree structure, which readily generalizes to 3D with an octree or other multi-way tree structures (Figure \ref{fig:amr_problems}). In this framework, the tokenizer partitions the input domain \( I \in \mathbb{R}^{H \times W \times c} \) based on specific physical criteria, transforming the structured grid into a set of patches \( I_p \in \mathbb{R}^{N \times K \times c} \), where each patch contains \( K = k \times k \) cells, represented as \( I_p = \text{AMRTokenizer}(I) \). The parameter \( k \) determines the number of subdivisions per dimension, directly influencing the resolution of the patches. 
To enhance computational efficiency, cells are processed in parallel at each depth before moving to the next, significantly accelerating the pruning process. The key steps of the AMR tokenizer are outlined as follows.
% This hierarchical representation preserves essential feature information at different depths, facilitating multi-scale analysis and reconstruction within the Transformer pipeline.
% \subsubsection{Parallel Processing Implementation}

% In the initialization phase, the entire input domain \( I \) is treated as a single patch for pruning. At each subsequent depth \( d \), regions resulting from the previous depth are further subdivided according to the pruning criteria. The maximum and minimum depths for subdivision are denoted by \( e \) and \( s \), respectively. Before reaching depth \( s \), all cells are subdivided but not stored. 

% For depths within the range \( s \leq d \leq e \), storage and subdivision decisions are made as follows: if a cell was marked as stored without subdivision in the previous layer, it is neither stored nor subdivided in the current layer. For all other cells, the \textbf{Navier-Stokes constraint-aware fast pruning module} evaluates pruning conditions concurrently. If a cell meets the subdivision criteria, it is both subdivided and stored; otherwise, it is stored without further subdivision. At the maximum depth \( d = s \), all cells are fully stored and are not subdivided further.

In the initialization phase, the entire input domain \( I \) is treated as a single patch with \( k \times k \) cells for storing or subdividing according to the pruning criteria. At next each subsequent depth, regions resulting from the previous depth are further processed. The process is outlined as follows:
1) If a cell was marked as not to be further subdivided in the previous layer, it is no longer processed. For all other cells, the \textbf{Navier-Stokes constraint-aware fast pruning module} evaluates pruning conditions concurrently. 
2) If a cell meets the subdivision criteria, then it is both stored  (Eq.\ref{eq:storage}) and subdivided in the quadtree structure. 
3) Otherwise, it is stored without further subdivision.
To reasonably control the granularity of subdivision, we define the minimum depth \( s \) and the maximum depth \( e \). Before reaching depth \( s \), all cells are subdivided but not stored. Upon reaching depth \( e \), subdivision ceases.

During pruning, the module identifies the positions requiring subdivision or storage at each depth \( d \). For each cell \( m_i \) (where \( i = 1, \ldots, M_d \) and \( M_d \) represents the number of cells that require storage), we compute an aggregated feature along with its positional information (depth \( d \) and mean coordinates \( (\bar{x}_{m_i}, \bar{y}_{m_i}) \)) as,
\begin{equation}
\label{eq:storage}
C_{m_i} = \left[ \frac{1}{|m_i|} \sum_{(x, y) \in m_i} I(x, y), \, d, \, \bar{x}_{m_i}, \, \bar{y}_{m_i} \right],
\end{equation}
where \( x \) and \( y \) denote the grid coordinates within \( m_i \). The information for each cell \( C_{m_i} \) is then added directly to the cumulative list \( C \), which collects all stored cell information across depths. % and cells.

The cumulative list \( C \) is then organized into patches by grouping adjacent cells based on their spatial positions. This transforms the structured grid into a set of patches \( I_p\). This hierarchical patch representation facilitates multi-scale analysis in the Transformer pipeline.

% This aggregation continues across depths, ensuring complete coverage of the domain once all cells have either reached the minimum size or have been transformed into tokens.

\begin{figure*}[htbp]
    \centering
    \includegraphics[width=0.98\textwidth]{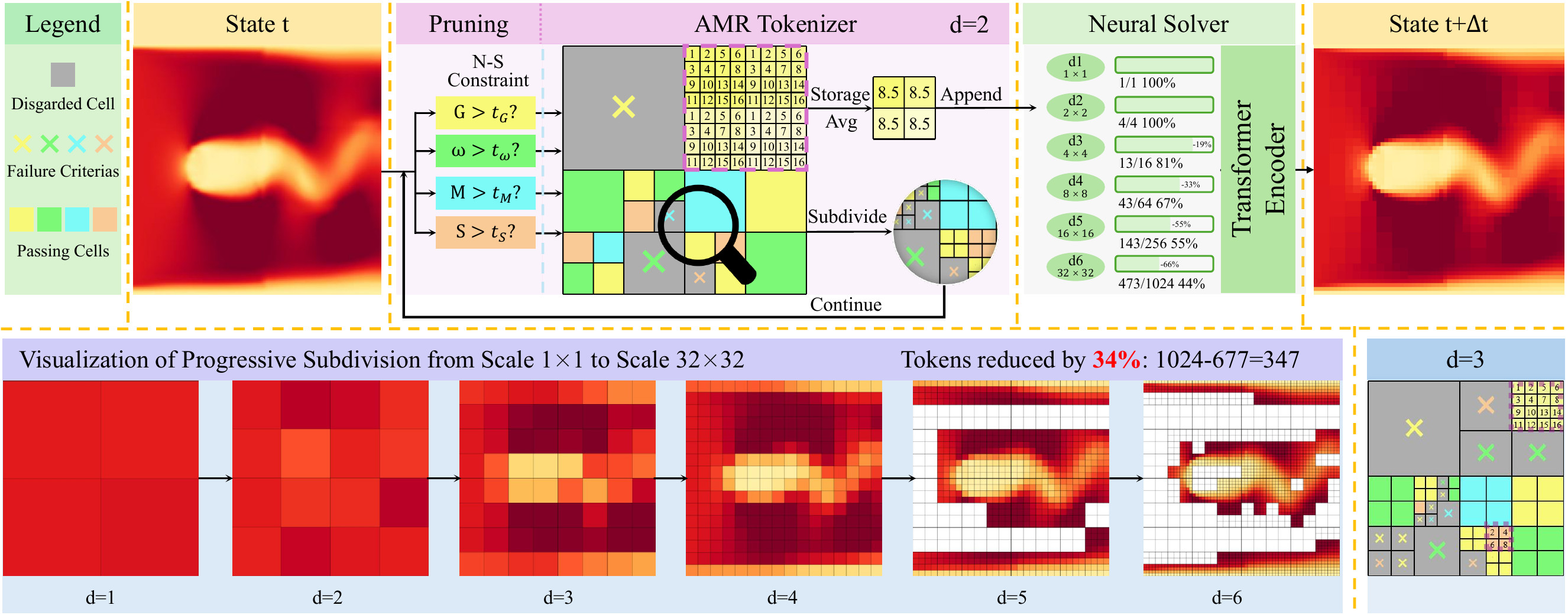}
    \caption{\textbf{AMR Tokenizer:} The AMR tokenizer adaptively partitions the input domain \( I \), where \( H \times W \times c \) represents the full spatial resolution of the domain with \( c \) channels of features. 
    AMR tokenizer refines the mesh progressively, from low resolution to high resolution, depth by depth.
    At each depth, cells undergo subdivision and storage based on customized Navier-Stokes constraints, including the velocity gradient, vorticity, momentum, and Kelvin-Helmholtz instability. 
    Grayscale cells marked with a colored × fail the corresponding physical constraint, meaning they are discarded and neither stored nor subdivided. Cells that pass the corresponding physical constraints are marked with the color of the most sensitive constraint, and stored and subdivided in the quadtree structure.
    The stored cells are aggregated by averaging (shown as "Storage (AVG)"), producing a compact multi-resolution representation of the domain. 
    The second row shows the mesh results of progressive subdivision at each depth.
    }
    \label{fig:tokenizer}
    \vspace{-0.4cm}
\end{figure*}

\subsection{N-S Constraint-Aware Pruning Module}
\label{sec:N-S Constraint-Aware Pruning Module}
The Navier-Stokes constraint-aware pruning module applies fast pruning based on specific physical properties that are crucial for capturing the complex dynamics of fluid flow. By concentrating computational resources in these regions, we ensure higher resolution where it matters most, thereby enhancing simulation accuracy. The criteria are designed for each cell \(m_i\) to capture unique aspects of fluid motion, ranging from localized rapid changes to instability-induced turbulence.

\noindent \textbf{Velocity Gradient:} The velocity gradient identifies sharp changes in fluid speed typically at boundaries, shock fronts, or regions with fine-scale features. For example, in shockwave and dam break scenarios, sudden velocity shifts create high gradients, indicating discontinuities or fronts. We compute the gradient magnitude \( G_{m_i} \) of cell \( m_i\) as,
\begin{equation}
G_{m_i} = \frac{1}{|m_i|} \int \sqrt{\nabla u^2 + \nabla v^2} \, dm_i,
% G = \sqrt{\left| \nabla u \right|^2 + \left| \nabla v \right|^2},
\end{equation}
where \( \nabla u \) and \( \nabla v \) represent spatial velocity changes. Capturing these shifts is critical for delineating features like shock fronts and complex flow boundaries, strengthening the model's tracking rapid transitions and flow separations.

\noindent \textbf{Vorticity:} The Vorticity \( \omega \) of cell \( m_i\) defined as,
\begin{equation}
\omega_{m_i} = \frac{1}{|m_i|} \int \left( \frac{\partial V}{\partial x} - \frac{\partial u}{\partial y} \right) \, dm_i,
% \omega = \frac{\partial v}{\partial x} - \frac{\partial u}{\partial y},
\end{equation}
measures local fluid rotation, indicating swirling flows like eddies or turbulence. High vorticity often characterizes regions where rotational motion drives turbulent mixing and energy dissipation, as seen in cavity flows or behind cylindrical obstacles. Identifying these zones ensures that vortex structures are resolved, supporting a detailed representation of rotational flow behavior crucial for understanding energy transfer within the fluid.

\noindent \textbf{Momentum:} The momentum identifies regions with significant flow that heavily influence the overall dynamics, particularly in cases where flow jets or moving currents, such as those in tube or dam flows, govern the system's behavior. The momentum \( M_{m_i} \) of cell \( m_i\) is defined as,  
%The momentum \( M \) of a region, calculated from the total components \( M_x \) and \( M_y \) as,
\begin{equation}
M_{m_i} = \frac{1}{|m_i|} \sqrt{\left( \int u \, dm_i \right)^2 + \left( \int v \, dm_i \right)^2},
% M = \sqrt{M_x^2 + M_y^2}, \quad M_x = \int u \, dA, \quad M_y = \int v \, dA,
\end{equation}
where \( A \) represents the area of the region being analyzed. Capturing momentum-heavy regions enables accurate simulation of flow-driving forces, impacting downstream behavior and the interaction of moving fronts with boundaries.

\noindent \textbf{Kelvin-Helmholtz Instability:} The Kelvin-Helmholtz instability detects shear-driven instabilities, which forms at fluid interfaces with significant velocity differentials. Such shear instabilities are common in atmospheric flows and can be relevant in tube and dam break scenarios, where layers of moving fluid interact. The shear strength \( S_{m_i}\) of cell \( m_i\) defined as,
\begin{equation}
S_{m_i} = \frac{1}{|m_i|} \int \left| \frac{\partial u}{\partial y} - \frac{\partial v}{\partial x} \right| \, dm_i,
% S = \left| \frac{\partial u}{\partial y} - \frac{\partial v}{\partial x} \right|,
\end{equation}
helps pinpoint areas prone to instability, allowing the model to depict the emergence of wave patterns and subsequent turbulence. This capability is essential for resolving the fine-scale interfacial interactions that contribute to mixing and layer formation.

We define a global set of characteristic physical properties \(\mathbf{P}_{\text{g}} = \{P_{G, \text{g}}, P_{\omega, \text{g}}, P_{M, \text{g}}, P_{S, \text{g}}\}\), where each property has an associated threshold factor \(\mathbf{T} = \{t_G, t_{\omega}, t_{M}, t_{S}\}\). The values of \(t_i\) are sampled uniformly over a predefined range, allowing manual adjustment of the AMR pruning parameters post-training to balance accuracy and efficiency. For a cell \( m_i \) with properties \(\mathbf{P}_{m_i} = \{P_{G, m_i}, P_{\omega, m_i}, P_{M, m_i}, P_{S, m_i}\}\), subdivision is triggered if,
\begin{equation}
\exists i \in \{G, \omega, M, S\} \quad \text{such that} \quad P_{i, m_i} > P_{i, \text{g}} \cdot t_i.
\end{equation}

For \textbf{Velocity Gradient} \( G \), we apply a proportional subdivision mechanism: if the velocity gradient in \( m_i \) at the current depth level falls within the top \( r_G \) percentile of the distribution at that level, then \( m_i \) undergoes subdivision if the following holds, where \( P_{G, d} \) denotes the distribution of velocity gradients at the current depth level and \(\text{Top-}r_G(P_{G, d})\) represents the top \( r_G \)-percentile value of \( P_{G, d} \), as,
\begin{equation}
P_{G, m_i} \geq \text{Top-}r_G\left(P_{G, d}\right).
\end{equation}

% where \( P_{G, d} \) represents the distribution of velocity gradients at the current depth level.

\subsection{Transformer Neural Solver}
\label{sec:neuralsolver}
% The objective of the neural solver is to solve the Navier-Stokes equations, which describe fluid flow dynamics, in order to predict the state of the flow field—such as velocity, pressure, and density (as applicable)—at the next time step \( t + \Delta t \) based on the flow field information at time \( t \). Specifically, the governing equations are defined as:
% \begin{equation}
% \frac{\partial \mathbf{u}}{\partial t} + (\mathbf{u} \cdot \nabla) \mathbf{u} = -\frac{1}{\rho} \nabla p + \nu \nabla^2 \mathbf{u},
% \end{equation}
% where \( \mathbf{u} = (u, v) \) represents the velocity field, \( p \) is the pressure, \( \rho \) is the fluid density, and \( \nu \) denotes the kinematic viscosity.

The neural solver aims to solve the Navier-Stokes equations, which describe fluid flow dynamics, to predict the state of the flow field, including the velocity field \( \mathbf{u} = (u, v) \), pressure \( p \), density field \( \rho \), and viscosity field \( \nu \) (as applicable), at the next time step \( t + \Delta t \), based on the corresponding information at \( t \) as,
\begin{equation}
\frac{\partial \mathbf{u}}{\partial t} + (\mathbf{u} \cdot \nabla) \mathbf{u} = -\frac{1}{\rho} \nabla p + \nu \nabla^2 \mathbf{u}.
\end{equation}

The input domain \( I \in \mathbb{R}^{H \times W \times c} \) is defined as:
\( I_{ij} = \{ u_{ij}, v_{ij}, \dots \} \)
where the primary components are the velocity fields \( u_{ij} \) and \( v_{ij} \). Additional channels may include other physical fields such as density \( \rho_{ij} \), viscosity \( \nu_{ij} \), and pressure \( p_{ij} \), as well as constant parameters that incorporate case-specific global attributes. At each time step \( t \), the input domain \( I_t \) is processed using an adaptive mesh refinement (AMR) tokenizer. This tokenizer generates a refined list of patches \( I_{p,t} \in \mathbb{R}^{N \times K \times (c+3)} \), where each patch contains aggregated features and positional encodings as,
% \vspace{-0.09cm}
\begin{equation}
I_{p,t} = \text{AMRTokenizer}(I_t).
\end{equation}
% \vspace{-0.09cm}
The Transformer architecture was chosen for the neural solver due to its capability to effectively capture long-range interactions through the self-attention mechanism. This feature is particularly advantageous for fluid dynamics simulations, which involve complex, cross-regional dependencies. Furthermore, the Transformer can seamlessly handle varying numbers of patches \( N \) without requiring adjustments for different input sizes, unlike cCNNs, which typically rely on fixed grid sizes, or GNNs, which require graph-based structures. This flexibility allows the Transformer to process all patches simultaneously, naturally adapting to the multi-scale representation from the AMR tokenizer. The Transformer-based neural solver then predicts the state at the next time step \( t + \Delta t \), yielding an output \( I_{p, t + \Delta t} \in \mathbb{R}^{N \times K \times c} \), as,
% \vspace{-0.09cm}
\begin{equation}
I_{p, t + \Delta t} = \text{Transformer}(I_{p, t}).
\end{equation}
% \vspace{-0.09cm}
During training, hyperparameters for the AMR subdivision criteria are randomly generated, allowing flexibility in balancing accuracy and efficiency. We use the normalized mean squared error (NMSE) as the loss function to ensure scale consistency across simulations. The training labels are processed through the AMR Tokenizer to create tokenized labels that match the multi-scale representation of the model’s output. During testing, the model's output \( I_{p,t + \Delta t} \) is mapped back onto the original grid for direct comparison with the ground truth.

To further refine the AMR tokenizer's pruning criteria, we define the velocity fields at both the current and previous time steps as \( \mathbf{u}_t \) and \( \mathbf{u}_{t-\Delta t} \). Using forward Euler integration, we estimate a virtual velocity field \( \mathbf{u}_{t+\Delta t}' \) as,
% \vspace{-0.09cm}
\begin{equation}
\mathbf{u}_{t+\Delta t}' = \mathbf{u}_t + (\mathbf{u}_t - \mathbf{u}_{t-\Delta t}),
\end{equation}
% \vspace{-0.09cm}
where \( \mathbf{u}_{t+\Delta t}' \) serves as an additional input for the AMR tokenizer’s refinement criteria. Both \( \mathbf{u}_t \) and \( \mathbf{u}_{t+\Delta t}' \) inform the AMR decision-making process, and the union of regions flagged for refinement based on either field defines the final set of regions for subdivision and storage. 
% The AMR tokenizer then applies this refined partition to the entire input field \( I \) at time step \( t \), subdividing and storing information as needed to ensure optimal spatial resolution across dynamically relevant regions.

% To improve the fidelity of the pruning conditions, we use the velocity fields from both the current and previous time steps, applying forward Euler integration to estimate a virtual velocity field for the next time step. The current and virtual velocity fields are then used as input for the pruning criteria, with their results combined to determine the final set of pruning regions.

% , specifically designed to capture long-range dependencies in 2D CFD problems. 
% Compared to traditional convolutional neural networks (CNNs), implicit neural representation (INR) models such as DeepONet, Neural Fourier Machines (NFM), and neural operators (NO) like Fourier Neural Operator (FNO), the Transformer architecture effectively leverages patches as tokens. Each patch token, represented as \(\mathbb{R}^{k \times D}\), encapsulates local fluid dynamics at an adaptively refined resolution, which is tailored to the complexity of the region. The self-attention mechanism in the Transformer enables efficient and accurate modeling of long-range dependencies across the simulation domain, facilitating multiscale information processing without incurring prohibitive computational costs.

% 符号表
\section{Experiment}
\label{sec:exper}
\begin{figure*}[t]
    \centering
    \includegraphics[width=\textwidth]{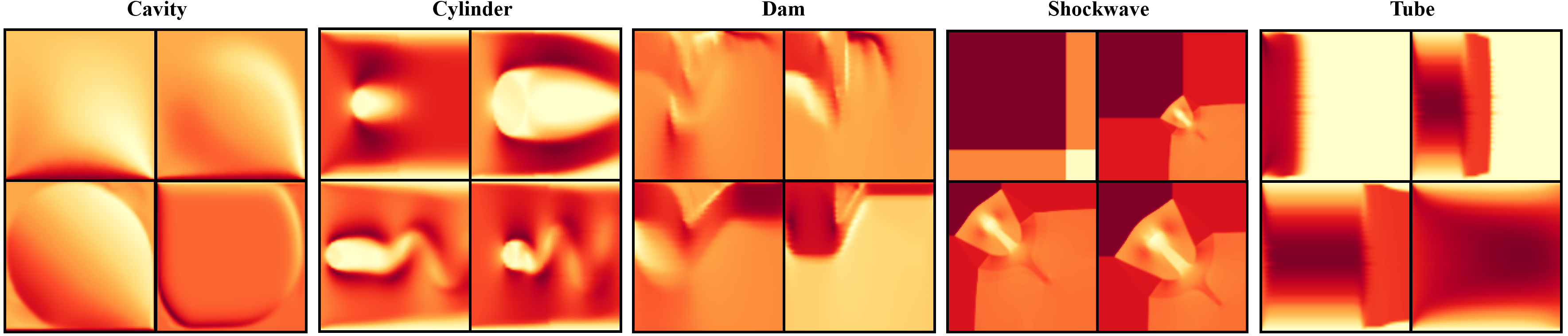}
    \caption{\textbf{Visualization of five problems:} Each column represents one of the five benchmark problems (Cavity, Cylinder, Dam, Shockwave, and Tube). These visualizations highlight the diversity and complexity of the flow scenarios, from boundary-driven flows in Cavity, vortex shedding around the Cylinder, rapid changes in the Dam break, steep gradients in Shockwave, to confined flows in the Tube.}
    \vspace{-0.4cm}
    \label{fig:amr_problems}
\end{figure*}

% This section evaluates the performance of our proposed AMR-Transformer on multiple fluid dynamics problems, including four benchmark problems from \texttt{CFDBench} (cylinder, dam, tube, and cavity) and a newly introduced shockwave dataset. 
We begin with the implementation details in Section~\ref{sec:implementation}, followed by an introduction to the \texttt{CFDBench} and \texttt{PDEBench} problems in Section~\ref{sec:cfdbench_cases} and a detailed analysis of our newly created compressible shockwave dataset in Section~\ref{sec:shockwave_analysis}. Comparative results between our model and current approaches are presented in Section~\ref{sec:sota_comparison}, with Section~\ref{sec:analysis_of_AMR} providing in-depth studies of the computational cost and accuracy impacts of the AMR tokenizer. Additional ablation studies are discussed in Section~\ref{sec:ablation}.

% \vspace{-0.1cm}
\subsection{Implementation}
\label{sec:implementation}
    Our Transformer architecture utilizes PyTorch's \texttt{nn.TransformerEncoder}~\cite{pytorch} as the neural solver, and the entire model is trained and evaluated on a single Nvidia RTX 4090 GPU. The number of epochs is set to 200 for the dam and shockwave problems, and to 500 for all other problems. For all problems, to adaptively refine regions based on physical properties, we set specific sampling ranges for each property. The threshold range for velocity gradient is uniformly sampled from 0.1 to 2, for momentum from 0.5 to 10, for vorticity from 0.2 to 4, and for Kelvin-Helmholtz(KH) instability from 0.2 to 4. We use a quadtree-based AMR structure with consistent model hyperparameters across tasks. The Transformer Neural Solver includes 4 attention heads, 6 transformer encoder layers, a hidden dimension of 256, and a feedforward network dimension of 1024. The batch size is set to 128, and optimization is performed with the Adam optimizer using a warmup-scheduled learning rate defined as,
    \begin{equation}
    \text{lr}(t) = \frac{1}{\sqrt{d_{\text{model}}}} \cdot \min \left( t^{-\frac{1}{2}}, \, t \cdot \textit{warmup\_steps}^{-\frac{3}{2}} \right),
    \end{equation}
    where \( d_{\text{model}} \) is set to 256, and \textit{warmup\_steps} is 4000.

    For the \texttt{CFDBench} dataset, the input domain \( I \) includes velocity components \( u \) and \( v \), along with global parameters like density and viscosity, with an output predicting velocity components at the next time step. For the \texttt{PDEBench} dataset, the input domain \( I \) is similar to the \texttt{CFDBench} dataset but without global parameters.
    For the shockwave dataset, \( I \) consists of \( u \), \( v \), pressure \( p \), and density \( \rho \), with output predictions for all four fields.

    % For the \texttt{CFDBench} dataset, the input domain \( I \in \mathbb{R}^{H \times W \times c_i} \) and patches \( P \in \mathbb{R}^{N \times k \times k \times c_i} \) have \( c_i=2+global\_case\_parameters \), where \( u \) and \( v \) represent the velocity components, and additional global parameters include characteristics like density and viscosity. The model’s output for \texttt{CFDBench} is the predicted \( u \) and \( v \) at the next time step, yielding an output dimension \( c_o=2 \).
    
    % For the shockwave dataset, \( c_i=4 \) channels are used, representing velocity components \( u \) and \( v \), pressure \( p \), and density \( \rho \). The model’s output \( c_o=4 \) includes predictions for all four fields, capturing the multi-physical nature of shockwave phenomena.

    % This schedule linearly increases the learning rate during the initial 4000 steps, after which it decays proportionally to the inverse square root of the training step.
    
    % Detailed information, including random range of AMR tokenizer's scaled threshold, network hyperparameters, input and output formats can be found in appendix \ref{sec:model_details}. 
    % Our datasets and code are publicly available at \href{https://github.com/Heiyanyan/Learning-Physical-Simulation-with-Message-Passing-Transformer}{https://github.com/Heiyanyan/Learning-Physical-Simulation-with-Message-Passing-Transformer}.

\subsection{\texttt{CFDBench} AND \texttt{PDEBench} Benchmark}
\label{sec:cfdbench_cases}
% 见fig 0
    We evaluate our model on the seven problems provided in \texttt{CFDBench} and \texttt{PDEBench}, and present the visualizations in Figure \ref{fig:amr_problems}:
    \begin{itemize}
        \item \textbf{Cylinder}: Simulates fluid flow around a stationary cylinder, capturing phenomena like flow separation, boundary layer behavior, and periodic vortex shedding. This problem has 185 cases with a total of 205,620 frames.
        \item \textbf{Dam}: Models the rapid release of water from a column collapse, representing free surface flows with complex interactions and varying flow velocities. It comprises a total of 220 cases and 21,916 frames.
        \item \textbf{Tube}: Examines a jet flow in a narrow tube, testing the model’s ability to simulate confined flow dynamics and the formation of boundary layers near the walls. There are 175 cases with 39,553 frames.
        \item \textbf{Cavity}: Represents driven cavity flow within a closed box, where a moving lid induces vortices and complex boundary-driven interactions, including a total of 159 cases and 34,582 frames.
        \item \textbf{Diffusion-Reaction}: Simulates the interaction of diffusion and chemical reactions, with substance concentrations evolving over time, including 1000 cases and 101,000 frames, with $128 \times 128$ grid.
        \item \textbf{NS-Incom-Inhom}: Simulates incompressible, inhomogeneous fluid flow using the Navier-Stokes equations with 4 cases and 4000 frames, with $512 \times 512$ grid.
    \end{itemize}
    These problems cover a variety of flow phenomena, from boundary layer effects and vortex shedding to free surface and confined flows, allowing for a broad evaluation of the AMR-Transformer's capabilities across fluid dynamics scenarios. However, all problems of CFDBench are generated at a low resolution of \(64 \times 64\), with a relatively uniform distribution of physical information across the domain.

    % More details of the benchmark can be found in Appendix \ref{sec:cfdbench_details}.
\vspace{-0.20cm}
\subsection{Shockwave Dataset}
\vspace{-0.1cm}
\label{sec:shockwave_analysis}
    To showcase our pipeline's capability in handling high-resolution simulations with uneven physical information distribution and capturing long-range dependencies, we introduce a new dataset, \textbf{Shockwave}, based on the 2D Riemann problem, specifically Configuration 3 from Kurganov and Tadmor~\cite{kurganov2002solution}. The dataset has a resolution of $128 \times 128$, four times larger than the $64 \times 64$ CFDBench datasets, and of comparable high resolution to the PDEbench dataset. Initial conditions are set as,
    \begin{equation}
    \resizebox{0.85\hsize}{!}{$
    \left(\rho, u, v, p \right) = \begin{cases}
    (1.5, 0, 0, 1.5), & \text{if} \quad x > 0.5, y > 0.5, \\
    (0.5323, 1.206, 0, 0.3), & \text{if} \quad x < 0.5, y > 0.5, \\
    (0.138, 1.206, 1.206, 0.029), & \text{if} \quad x < 0.5, y < 0.5, \\
    (0.5323, 0, 1.206, 0.3). & \text{if} \quad x > 0.5, y < 0.5.
    \end{cases}
    $}
    \end{equation}
    Here, $\rho$ represents density, $u$ and $v$ are velocity components in the $x$ and $y$ directions, and $p$ is pressure. The computational domain is $[0,1]^2$ with Neumann boundary conditions, and simulations are run to a final time of $t = 0.3$. To increase variability, random perturbations (up to $20\%$ deviation) are applied to initial conditions, creating 10 unique cases, each with 200 frames and containing $u$, $v$, $p$, and $\rho$ for detailed flow representation.

    Compared to CFDBench cases like cylinder and cavity flows, which involve relatively simple, steady-state or periodic structures, the Shockwave dataset introduces strong shocks, sharp discontinuities, and complex, evolving small-scale vortex structures, creating a challenging testbed for models to capture intricate fluid dynamics.

    % This problem is uniquely suited to evaluating the performance of the proposed AMR-Transformer framework, as it features highly uneven distributions of physical information that challenge the efficiency of AMR. Specifically, the AMR-Transformer allows us to assess its capability to accurately identify regions with high gradients for grid refinement, thereby increasing local accuracy without excessive computational overhead. In contrast, simple regular tokenization leads to significant computational redundancy when handling such heterogeneous information distributions.
    % More detailed intruduction can be found in Appendix \ref{sec:shockwave_appendix}.

\subsection{Comparative Analysis with SOTA}
\label{sec:sota_comparison}
% \begin{table}[ht]
% \centering
% \caption{Comparative Analysis of AMR-Transformer with SOTA Models Across Five Problems}
% \label{tab:sota_comparison}
% \begin{tabular}{|c|c|c|c|c|}
% \hline
% \textbf{Problem} & \textbf{Model} & \textbf{NMSE} & \textbf{MAE} & \textbf{MSE} \\
% \hline
% \multirow{2}{*}{Dam} & SOTA & 2.7361E-3 & 7.03E-3 & 1.49E-3 \\
%                      & Ours & \textbf{4.27E-4} & \textbf{3.66E-3} & \textbf{1.63E-4} \\
% \hline
% \multirow{2}{*}{Cylinder} & SOTA & \textbf{1.78E-5} & 3.06E-3 & 2.74E-5 \\
%                           & Ours & 4.12E-5 & \textbf{2.24E-3} & \textbf{2.09E-5} \\
% \hline
% \multirow{2}{*}{Cavity} & SOTA & \textbf{4.166E-4} & 3.19E-2 & 1.58E-2 \\
%                         & Ours & 5.76E-4 & \textbf{2.47E-2} & \textbf{4.71E-3} \\
% \hline
% \multirow{2}{*}{Tube} & SOTA & \textbf{3.19E-3} & 1.82E-2 & \textbf{1.27E-3} \\
%                       & Ours & 3.49E-3 & \textbf{1.28E-2} & 1.98E-3 \\
% \hline
% \multirow{2}{*}{Shock} & SOTA & 5.53E-3 & 3.53E-2 & 4.83E-3 \\
%                        & Ours & \textbf{9.32E-4} & \textbf{1.71E-2} & \textbf{9.66E-4} \\
% \hline
% \end{tabular}
% \end{table}

    \begin{table}[!htbp]
    \vspace{-0.3cm}
    \centering
    \resizebox{0.45\textwidth}{!}{
    \begin{tabular}{c c c c c}
    % \hline
    \toprule
    \textbf{Problem} & \textbf{Model} & \textbf{NMSE} & \textbf{MAE} & \textbf{MSE} \\
    % \hline
    \midrule
    \multirow{5}{*}{Dam} & Identity & 3.16E-3 & 8.20E-3 & 1.69E-3 \\
                         & U-Net   & 3.24E-3 & 9.21E-3 & 1.70E-3 \\
                         & FNO     & 6.36E-2 & 1.02E-1 & 2.03E-2 \\
                         & DeepONet  & 3.08E-3 & 7.27E-3 & 1.64E-3 \\
                         & \textbf{Ours}    & \textbf{4.10E-4} & \textbf{3.24E-3} & \textbf{1.63E-4} \\
    % \hline
    \midrule
    \multirow{5}{*}{Cylinder} & Identity & 1.57E-2 & 1.09E-1 & 7.54E-2 \\
                              & U-Net    & 2.16E-5 & 3.09E-3 & 5.49E-5 \\
                              & FNO      & \textbf{1.78E-5} & 3.06E-3 & 2.74E-5 \\
                              & DeepONet & 6.86E-2 & 1.27E-1 & 5.43E-2 \\
                              & \textbf{Ours}     & 4.12E-5 & \textbf{2.24E-3} & \textbf{2.09E-5} \\
    % \hline
    \midrule
    \multirow{5}{*}{Cavity} & Identity & 1.41E-3 & 4.11E-1 & 5.95E-1 \\
                            & U-Net    & \textbf{4.166E-4} & 3.19E-2 & 1.58E-2 \\
                            & FNO      & 5.06E-4 & 5.69E-2 & 1.77E-2 \\
                            & DeepONet & 1.39E-3 & 5.66E-2 & 6.38E-2 \\
                            & \textbf{Ours}     & 5.76E-4 & \textbf{2.47E-2} & \textbf{4.71E-3} \\
    % \hline
    \midrule
    \multirow{5}{*}{Tube} & Identity & 1.11E-1 & 1.20E-1 & 1.64E-1 \\
                          & U-Net    & \textbf{3.18E-3} & 1.81E-2 & 1.29E-3 \\
                          & FNO      & 5.30E-3 & 2.39E-2 & \textbf{1.27E-3} \\
                          & DeepONet & 6.48E-2 & 1.20E-1 & 7.23E-2 \\
                          & \textbf{Ours}     & 3.49E-3 & \textbf{1.13E-2} & 1.33E-3 \\
    % \hline
    \midrule
    \multirow{5}{*}{Shock} & Identity & 7.42E-1 & 4.11E-1 & 5.95E-1 \\
                           & U-Net    & 7.20E-2 & 1.59E-1 & 6.21E-2 \\
                           & FNO      & 5.53E-3 & 3.53E-2 & 4.83E-3 \\
                           & DeepONet & 6.86E-2 & 1.27E-1 & 5.43E-2 \\
                           & \textbf{Ours}     & \textbf{9.32E-4} & \textbf{1.71E-2} & \textbf{9.66E-4} \\
    \midrule
    \multirow{5}{*}{Diffusion-Reaction} & Identity & 3.11E+1 & 8.07E-1 & 1.02 \\
                           & U-Net    & 4.72E+1 & 8.04E-1 & 1.01 \\
                           & FNO      & \textbf{8.14E-1} & 1.03E-1 & 1.76E-2 \\
                           & DeepONet & 3.09E+1 & 8.05E-1 & 1.01 \\
                           & \textbf{Ours}     & 8.17E-1 & \textbf{1.02E-1} & \textbf{1.75E-2} \\
    \midrule
    \multirow{5}{*}{NS-Incom-Inhom} & Identity & 1.49E-3 & 1.47E-3 & 5.61E-4 \\
                           & U-Net    & 1.23E-3 & 2.48E-3 & 1.50E-5 \\
                           & FNO      & 1.00 & 1.14E-1 & 4.24E-2 \\
                           & DeepONet & 5.68E-5 & 8.96E-4 & 5.88E-6 \\
                           & \textbf{Ours}     & \textbf{5.42E-5} & \textbf{5.87-4} & \textbf{3.76E-6} \\
    % \hline
    \bottomrule
    \end{tabular}
    }
    \caption{\textbf{Performance comparison across models on seven problems.} The "Identity" model serves as a baseline reference, returning the input as output without any learned modifications.}
    \label{tab:model_comparison}
    \vspace{-0.2cm}
    \end{table}
    
    We compare the AMR-Transformer with current state-of-the-art (SOTA) models, including U-Net~\cite{unet}, FNO~\cite{fno}, and DeepONet~\cite{deeponet}, across all seven problems (cylinder, dam, tube, cavity, shockwave, diffusion-reaction and ns-incom-inhom). U-Net and FNO are CNN-based architectures, while DeepONet represents an implicit neural representation approach. The results in Table~\ref{tab:model_comparison} reveal significant improvements for our model. Model-specific hyperparameters follow the configurations in \texttt{CFDBench}~\cite{cfdbench}, ensuring consistency across comparisons.
    
    Our AMR-Transformer pipeline consistently displays strong learning capabilities across all problems, with particularly notable improvements in the most challenging scenarios. 
    % The pipeline demonstrated the most significant improvements in accuracy on the Dam and Shock problems, outperforming SOTA models by nearly \textbf{10 times} on key metrics, achieving an order of magnitude improvement in accuracy. 
    For the Dam problem, our model achieves a relative improvement of approximately \textbf{91\%} on the MSE metric, reducing it to 1.63E-4 compared to DeepONet's 1.64E-3. This problem involves complex flow interactions and rapid changes that demand a model capable of handling non-local features, which our pipeline effectively addresses. Similarly, in the Shock problem, characterized by steep gradients and abrupt transitions, the AMR-Transformer surpasses the best alternative, FNO, by approximately \textbf{83\%} on NMSE
    , showcasing its robust handling of multiscale phenomena.
    % , achieving 9.32E-4 compared to FNO's 5.53E-2
    
    % In the Cylinder and Cavity problems, our model achieved competitive results, ranking among the best across multiple evaluation metrics. 
    In the Cylinder problem, our model’s MAE is reduced by approximately \textbf{27\%} compared to FNO, underscoring the pipeline's effectiveness in capturing flow separations, reattachments, and eddy-driven energy transfers. Notably, the Cylinder problem showcases periodic behaviors, which FNO excels at due to its proficiency in learning repetitive structures. In the Cavity problem, our model demonstrates a substantial improvement in MSE, achieving a reduction of \textbf{70\%} compared to FNO, highlighting our model’s precision in capturing boundary-driven interactions and vortical structures within enclosed flow domains.
    
    In the Dam and NS-Incom-Inhom problems, the FNO's MAE is 12.4 times and 75.6 times higher than the identity baseline, respectively, clearly indicating its inability to learn the underlying dynamics of this complex scenario. Similarly, DeepONet struggles on both Cylinder and Dam problems, with performance metrics close to the Identity baseline, reflecting its limitations in handling high-gradient and complex conditions. 

    In the Diffusion-Reaction problem, U-Net and DeepONet fail to capture the complex variations. In contrast, both FNO and our model effectively handle these challenges. In the NS-Incom-Inhom problem, characterized by complex fluid dynamics and inhomogeneous properties, our model outperforms DeepONet by approximately \textbf{36\%} on MSE.
    % , demonstrating its superior ability to handle the challenging flow characteristics.

    % Our AMR-Transformer pipeline demonstrated consistently high performance across all problems, with the most notable accuracy improvements on the challenging Dam and Shock problems, achieving an order of magnitude improvement over SOTA models. Specifically, our model outperformed DeepONet on the Dam problem by approximately \textbf{91\%} in MSE, effectively handling complex flow interactions and rapid changes. For the Shock problem, characterized by steep gradients, the AMR-Transformer surpassed FNO by \textbf{83\%} in NMSE, highlighting its strong multiscale processing capability.

    % In the Cylinder and Cavity problems, our model also performed competitively. For the Cylinder problem, it reduced MAE by \textbf{27\%} compared to FNO, capturing flow separations and eddy-driven energy transfers effectively. The model's MSE improvement of \textbf{70\%} on the Cavity problem demonstrates its precision in boundary-driven interactions.

    % While our model excelled, FNO and DeepONet exhibited limitations. FNO’s MAE on the Dam problem was \textbf{12.4 times} higher than the Identity baseline, indicating difficulty in learning complex dynamics. DeepONet performed near the Identity baseline on both Cylinder and Dam problems, and U-Net struggled with the Shock problem, showing up to \textbf{99\%} higher NMSE than our approach.
    
    Based on the results in \texttt{CFDBench}~\cite{cfdbench} and \texttt{PDEBench}~\cite{PDEBENCH}, we exclude Physics-Informed Neural Networks~\cite{pinn} from this comparison, as they did not exceed baseline performance in any tested scenario.
    
\subsection{Analysis of AMR tokenizer}
    \vspace{-0.1cm}
    \label{sec:analysis_of_AMR}
    To analyze the impact of the AMR tokenizer, we examine its influence on computational cost and accuracy in Table \ref{tab:N-S Constraint-Aware Pruning Module ablation}.
    % \vspace{-0.3cm}
    
    \begin{table}[h!]
    \vspace{-0.2cm}
        \centering
        \resizebox{0.45\textwidth}{!}{
        \begin{tabular}{l l c c}
        % \hline
        \toprule
        \textbf{Dataset}   & \textbf{Method}  & \multicolumn{1}{c}{\begin{tabular}[c]{@{}c@{}}\textbf{Token Count(N)}\end{tabular}} & \multicolumn{1}{c}{\begin{tabular}[c]{@{}c@{}}\textbf{GFLOPs}\end{tabular}} \\
        % \hline
        \midrule
        \multirow{2}{*}{Shock}              & Regular          & \(4096\)          & 71.37                    \\
                      & AMR              & \(970_{\pm 388}\)          & 7.51                     \\
        % \hline
        \midrule
        \multirow{2}{*}{Diffusion-Reaction} & Regular & \(4096\) & 71.37 \\
                           & AMR    & \(1296_{\pm 634}\) & 11.34 \\
        \midrule
        \multirow{2}{*}{NS-Incom-Inhom} & Regular & \(65536\) & 13607.90 \\
                           & AMR    & \(7547_{\pm 4252}\) & 212.12 \\
        \midrule
        CFDBench           & Regular          & \(1024\)           & 8.10                     \\
        Cylinder           & AMR              & \(354_{\pm 212}\)          & 2.07                     \\
        Cavity             & AMR              & \(509_{\pm 367}\)          & 3.22                     \\
        Dam                & AMR              & \(347_{\pm 210}\)          & 2.02                     \\
        Tube               & AMR              & \(498_{\pm 295}\)          & 3.13                     \\
        % \hline
        \bottomrule
        \end{tabular}
        }
        \caption{\textbf{Computational cost comparison}: The table shows the average number of tokens generated (with standard deviation) and the computational cost in gigaflops (GFLOPs) for each dataset and method. "CFDBench-Regular" represents four problems simulated with a regular grid, all having identical computational efficiency.}
        \label{tab:flops_token_count}
        \vspace{-0.3cm}
    \end{table}
Given the \(O(N^2)\) complexity of self-attention, reducing the token count results in quadratic reductions in FLOPs. For the shockwave dataset, the AMR tokenizer improves computational efficiency by nearly \textbf{10 times} by significantly decreasing both token count and FLOPs compared to the regular grid (Figure \ref{tab:flops_token_count}). While the regular tokenizer maintains a fixed token count and FLOPs across CFDBench datasets, the AMR tokenizer adapts the token count based on each problem's complexity, achieving approximately \textbf{60\% to 70\%} reductions in tokens and \textbf{70\% to 75\%} reductions in FLOPs. For the PDEBench dataset, the AMR tokenizer achieves reductions of \textbf{61\% to 88\%} in tokens, and \textbf{84\% to 98\%} in FLOPs. This adaptive refinement enables the AMR tokenizer to focus computational resources on dynamically significant regions, ensuring high efficiency.

To quantitatively assess the AMR tokenizer’s performance, we apply it to a 1024 × 1024 dataset containing velocity fields for shock wave and explosion simulations, aimed at generating visually accurate effects (Figure \ref{fig:amr_tokenizer_results}). The AMR tokenizer achieved a \textbf{4 to 10 times} reduction in token count compared to a 512 × 512 regular grid while preserving essential details. Furthermore, it significantly improves accuracy, reducing MSE by factors ranging from \textbf{6 to 1000} over the regular grid. These results highlight the tokenizer’s ability to balance computational efficiency with high fidelity in capturing critical features, demonstrating its adaptability for high-resolution physical simulations and potential applications in visual effect generation.
% \vspace{-0.1cm}

\subsection{Ablation Study}
\vspace{-0.1cm}
\label{sec:ablation}
\begin{table}[ht]
    % \vspace{-0.4cm}
    \centering
    \resizebox{0.4\textwidth}{!}{
    \begin{tabular}{l c c}
    \toprule
    \textbf{Physical Property} & \textbf{Token Pruned} & \textbf{MSE} \\ 
    \midrule
    Velocity Gradient & 257 & 2.80E-4 \\
    Vorticity & 80 & 1.93E-3 \\
    Momentum & 132 & 1.12E-3 \\
    KH Instability & 187 & 5.40E-4 \\
    \midrule
    Overall & 347 & 1.63E-4 \\
    Regular & 1024 & 1.36E-4 \\
    \bottomrule
    \end{tabular}
    }
    \caption{\textbf{N-S Constraint-Aware Pruning Module ablation}: Presentation of the average number of tokens and MSE values for applying different physical properties in the dam problem. The "Regular" row represents the results using a standard, non-adaptive tokenizer without N-S constraint-aware pruning.}
    \label{tab:N-S Constraint-Aware Pruning Module ablation}
    \vspace{-0.6cm}
\end{table}

The ablation study in Table \ref{tab:N-S Constraint-Aware Pruning Module ablation} demonstrates that using individual physical properties alone does not achieve optimal results in terms of accuracy. In contrast, the "Overall" method, which incorporates all physical properties, markedly enhances accuracy to an MSE of \(1.63 \times 10^{-4}\) while requiring a moderate increase in token number (347). This result highlights that integrating multiple criteria allows the model to capture a broader spectrum of dynamic flow features, leading to improved precision with only a slight increase in computational cost. Compared to the standard, non-adaptive regular tokenizer, which produces an MSE of \(1.36 \times 10^{-4}\) with 1024 tokens, our AMR tokenizer achieves a similar simulation accuracy (1.63E-4) while using much fewer tokens, illustrating the efficiency of constraint-aware adaptive refinement.

\begin{table}[ht]
    \vspace{-0.1cm}
    \centering
    \resizebox{0.38\textwidth}{!}{
    \begin{tabular}{l c c c}
    \toprule
    \textbf{Neural Solver} & \textbf{MSE} & \textbf{MAE} & \textbf{NMSE} \\
    \midrule
    AMR-Transformer & \textbf{9.32E-4} & \textbf{1.71E-2} & \textbf{9.66E-4} \\
    AMR-MeshGraphNet & 7.38E-2 & 1.40E-1 & 8.16E-2 \\
    \bottomrule
    \end{tabular}}
    \caption{\textbf{Neural Solver comparison}: Performance comparison of the Transformer and MeshGraphNet on the shockwave problem.}
    \label{tab:neural_solver_comparison}
    \vspace{-0.25cm}
\end{table}

We conduct an ablation study replacing our Transformer with MeshGraphNet \cite{mgn} in Table \ref{tab:neural_solver_comparison}, configured with a hidden layer dimension of 128, 15 message-passing steps, and edges connecting nodes within a radius of 4. The results demonstrate the Transformer’s superior accuracy, underscoring its effectiveness in capturing the long-range interaction of the shockwave problem.
\vspace{-0.1cm}

\section{Conclusion}
\label{sec:conclu}
\vspace{-0.1cm}
We introduced AMR-Transformer, a novel pipeline that leverages Adaptive Mesh Refinement with an Encoder-only Transformer to address the challenges of complex fluid dynamics simulations. Our approach adaptively allocates computational resources to the most complex regions while utilizing the Transformer’s self-attention mechanism to capture global interactions across scales. This combination achieves an excellent balance between efficiency and accuracy, making the model well-suited for scenarios that require both detailed spatial resolution and effective global context modeling.

\section{Acknowledgment} 
\vspace{-0.1cm}
This work was supported by National Science Foundation of China (U20B2072, 61976137). This work was also partially supported by Grant YG2021ZD18 from Shanghai Jiaotong University Medical Engineering Cross Research. This work was partially supported by STCSM 22DZ2229005.

% Despite its promise, the current evaluation is limited by the resolution and scope of available datasets, which often lack the multiscale features necessary to fully showcase the strengths of our method. Additionally, we employed a basic Encoder-only Transformer as the neural solver; exploring more advanced architectures could further enhance the model’s capability to capture intricate fluid behaviors. Future work will focus on expanding the AMR-Transformer to more sophisticated Transformer designs and testing it on richer, multiscale datasets to fully realize its potential in complex simulations.

{
    \small
    \bibliographystyle{ieeenat_fullname}
    \bibliography{main}
}
% \appendix
% \input{texts/Appendix}

% \input{texts/Intro}
% \input{texts/Intro}
% \input{sec/0_abstract}    
% \input{sec/1_intro}
% \input{sec/2_formatting}
% \input{sec/3_finalcopy}


\end{document}